\documentclass[letterpaper, 10 pt, conference]{ieeeconf}  
\usepackage{tikz}
\usetikzlibrary{arrows,decorations.pathreplacing}
\usepackage{booktabs}
\newcommand*\rot{\rotatebox{50}}
\usepackage{amsmath}
\usepackage{subcaption}
\usepackage{multirow}
\usepackage{amssymb}
\usepackage{pifont}
\usepackage{adjustbox}
\usepackage{hyperref}

\usepackage{epstopdf}
\newcommand{\cmark}{\ding{51}}%

\IEEEoverridecommandlockouts                              

\title{\LARGE \bf
Detecting Traffic Lights by Single Shot Detection
}

\author{Julian M\"uller$^{1}$ and Klaus Dietmayer$^{1}$
\thanks{$^{1}$Julian M\"uller and  Klaus Dietmayer are with Department of Measurement, Control and Microtechnology, Ulm University, 89081 Ulm, Germany
        {\tt\small \{julian-2.mueller, klaus.dietmayer\}@uni-ulm.de}}%
}

\newcommand{\etal}{\textit{et al}.}

\begin{document}

\maketitle
\thispagestyle{empty}
\pagestyle{empty}

\begin{abstract}
Recent improvements in object detection are driven by the success of convolutional neural networks (CNN). They are able to learn rich features outperforming hand-crafted features. So far, research in traffic light detection mainly focused on hand-crafted features, such as color, shape or brightness of the traffic light bulb.\\
This paper presents a deep learning approach for accurate traffic light detection in adapting a single shot detection (SSD) approach. SSD performs object proposals creation and classification using a single CNN. The original SSD struggles in detecting very small objects, which is essential for traffic light detection. By our adaptations it is possible to detect objects much smaller than ten pixels without increasing the input image size. We present an extensive evaluation on the DriveU Traffic Light Dataset (DTLD). We reach both, high accuracy and low false positive rates. The trained model is real-time capable with ten frames per second on a Nvidia Titan Xp.
Code has been made available at \url{https://github.com/julimueller/tl_ssd}. 
\end{abstract}


%
\IEEEpeerreviewmaketitle

\section{Introduction}

Traffic light detection is a key problem for autonomous driving. The basis publication in traffic light detection forms Lindner \etal~\cite{1336354}, using color and shape for proposal generation.
This conventional proceed - creating proposals (or also called candidates) by hand-selected features and a subsequent verification/classification - characterized publications in the following decade.\\
A main drawback of separating the object proposal generation step and classification is runtime and accuracy. Popular region proposal algorithms are not real-time capable~\cite{Uijlings13,edge-boxes-locating-object-proposals-from-edges} often requiring several seconds per image. In addition, hand-crafted features for traffic light detection are not able to reach sufficient accuracy~\cite{8317948} saturating clearly below 100 percent.\\
With rising success of CNNs, also object proposal generation was performed by sharing a base network together with classification~\cite{Ren2015, Liu2015}. Typically, those networks are trained by a confidence and localization loss~\cite{Erhan2014}, guaranteeing accurate bounding boxes with respect to the overlap metric intersection over union. One key problem of those approaches is the detection of small objects. It is mainly caused by pooling operations, which increase the receptive field and reduce the computational effort. However, pooling also decreases the image resolution leading to difficulties for accurate localization of small objects. 
\begin{figure}[!t]
\centering
\includegraphics[width=\linewidth]{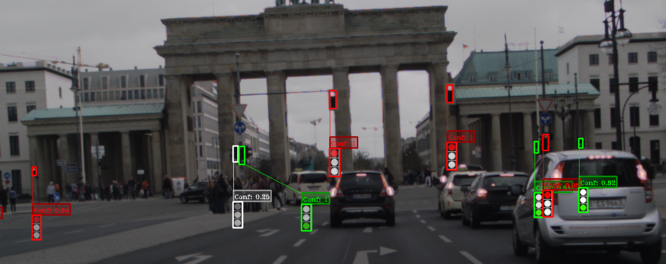}
\caption{Exemplary results of our SSD approach for traffic light detection. Our approach is able to even detect objects smaller or equal five pixels in width and moreover predicts their state by an additional branch. }
\end{figure}
\\In this paper, we present the \emph{TL-SSD}, an adaption of the original \emph{Single Shot Detector} (SSD)~\cite{Liu2015}, trained on the large-scale \emph{DriveU Traffic Light Dataset}~\cite{DBLP:conf/icra/Fregin}.
We make the following contributions:
\begin{enumerate}
\item We prove that the original approach struggles in detecting small objects in later feature layers due to a prior box stride based on the layer size. We demonstrate a possibility how to detect small objects in later layers without subsampling the layer itself.
\item We replace the original base network by an Inception network~\cite{DBLP:journals/corr/SzegedyVISW15}, which is faster and more accurate
\item We extend the approach for state (color) prediction by adding a convolutional layer and extending loss calculation
\item We adapt the class-wise non maximum suppression to a class-independent one to avoid multi-detections.
\item We present an extensive evaluation on a large-scale dataset with very promising results
\end{enumerate}  
We organize our paper as follows: The next section briefly describes the state-of-the-art. We differentiate classical traffic lights detection methods from CNN-based methods. The TL-SSD explaining the original approach and all modifications is presented in Section~\ref{sec:tlssd}. Section~\ref{sec:experiments} presents extensive experiments on the DTLD~\cite{DBLP:conf/icra/Fregin}. The paper closes with a conclusion in Section~\ref{sec:conclusion}.

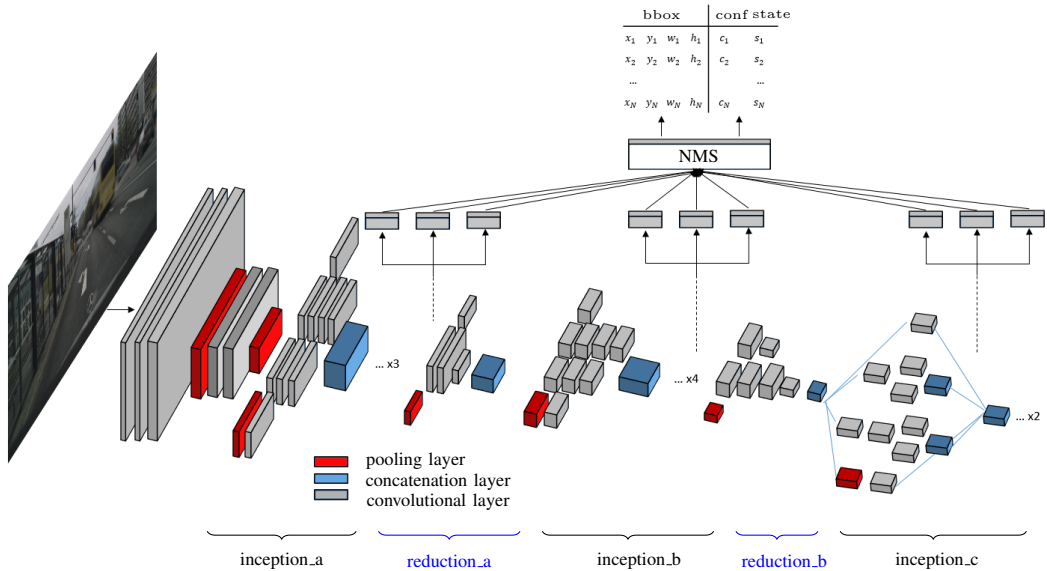
\begin{figure*}[!t]
	\centering
	\scalebox{0.99}{
	\begin{tikzpicture}
			\node[inner sep=0pt] (russell) at (0,0)
		  {\includegraphics[width=0.8 		\linewidth]{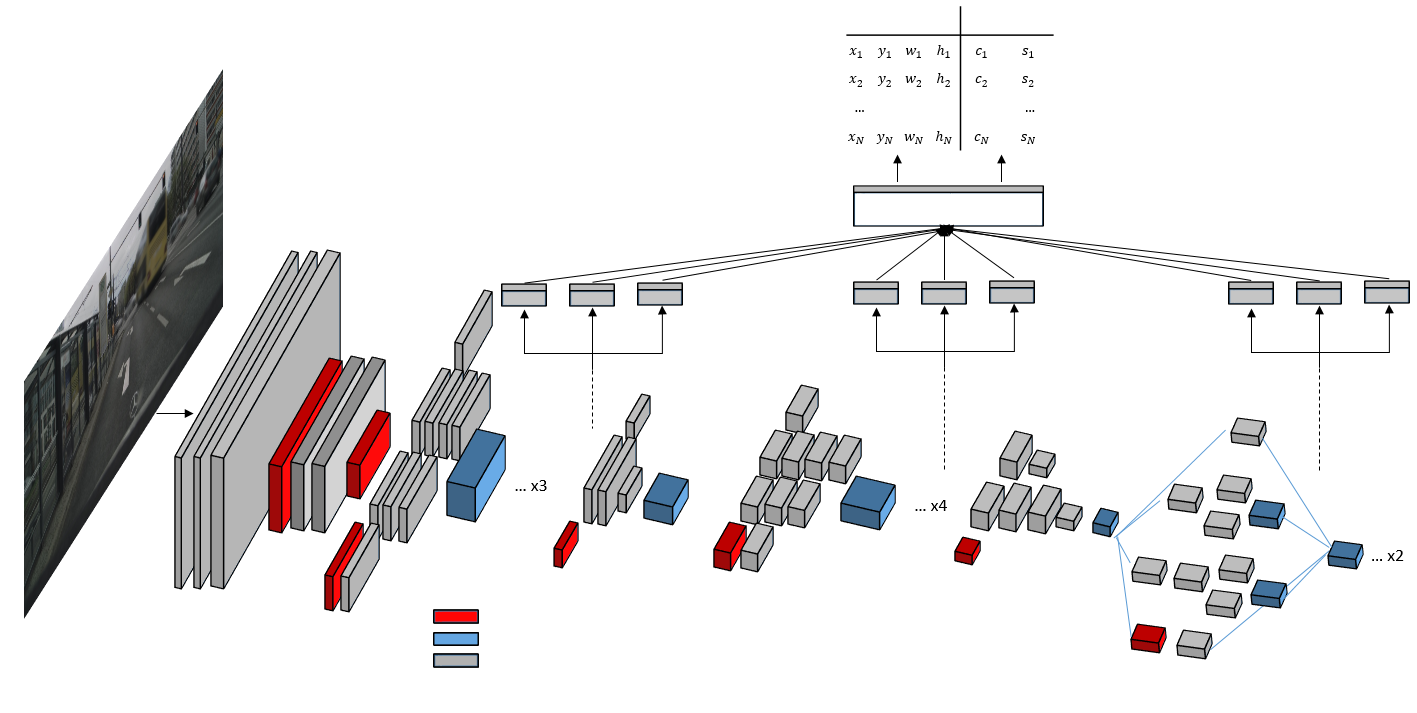}};
		 
			\node[] at (1.9,3.4) {\tiny $\mathrm{bbox}$};
			\node[] at (2.85,3.4) {\tiny $\mathrm{conf}$};
			\node[] at (3.38,3.4) {\tiny $\mathrm{state}$};
			\node[xshift = 2.6cm, xshift = 0.8cm, yshift = -0.8cm] 	at (-4.8,-1.8) {\scriptsize pooling layer};
			\node[xshift = 2.6cm, xshift = 0.8cm, yshift = -0.8cm] 	at (-4.5,-2.08) {\scriptsize concatenation layer};
			\node[xshift = 2.6cm, xshift = 0.8cm, yshift = -0.8cm] 	at (-4.5,-2.35) {\scriptsize convolutional layer};
			\node[] at (2.4,1.5) {\scriptsize NMS};
			\draw[decorate,decoration={brace,amplitude=3pt,mirror}, 	xshift = 0.8cm, yshift = -0.55cm] 
		  (3.5,-3)  -- (6,-3);
			\draw[decorate,decoration={brace,amplitude=3pt,mirror},, 	xshift = 0.8cm, yshift = -0.55cm, blue] 
		  (2.1,-3)  -- (3.2,-3);
			\draw[decorate,decoration={brace,amplitude=3pt,mirror},, 	xshift = 0.8cm, yshift = -0.55cm, blue] 
		  (-2.7,-3)  -- (-0.8,-3);
			\draw[decorate,decoration={brace,amplitude=3pt,mirror}, 	xshift = 0.8cm, yshift = -0.55cm] 
		  (-0.5,-3)  -- (1.8,-3);  
			\draw[decorate,decoration={brace,amplitude=3pt,mirror}, 	xshift = 0.8cm, yshift = -0.55cm] 
			(-5,-3)  -- (-3,-3);   
			\node[, xshift = 0.1cm, yshift = -0.55cm] at (5.5,-3.4) 	{\scriptsize inception\_c};
			\node[, xshift = 0.8cm, yshift = -0.55cm] at (.8,-3.4) 	{\scriptsize inception\_b};
			\node[, xshift = 0.8cm, yshift = -0.55cm] at (-4,-3.4) 	{\scriptsize inception\_a};
			\node[blue, xshift = 0.55cm, yshift = -0.55cm] at 	(3.0,-3.4) {\scriptsize reduction\_b};
			\node[blue, xshift = 0.8cm, yshift = -0.55cm] at 	(-1.75,-3.4) {\scriptsize reduction\_a};
	\end{tikzpicture} }              
	\caption{Single Shot Detection with an Inception\_v3 base network. The input image is of size 2048x512 pixels. There are five convolutional and two pooling layers in the beginning. Three Inception\_A layers follow, which consist of several convolutional layers with filters of different size. Reduction layers reduce the layer size. Afterwards, four Inception\_B layers, a reduction layer and two Inception\_C layers follow. Prior boxes are created on top of several convolutional layers. Bounding boxes are predicted from additional convolutional layers, which predict offsets with respect to the prior boxes. Confidence and state are predicted from another convolutional layer. A non-maximum suppression is performed on the resulting predictions.}
	\label{fig:ssd_net}
\end{figure*}
\section{Related Work}
We separate the related work in three different parts. First of all, approaches generating object proposals from ConvNets are presented. Afterwards we briefly present publications on traffic light detection and approaches using CNNs in detail.\\
\textbf{Object proposals from CNNs:}
OverFeat~\cite{Sermanet2013} detects bounding box coordinates from a fully-connected layer. A more general approach is MultiBox~\cite{Erhan2014} predicting bounding boxes for multi-class tasks from a fully-connected layer. YOLO~\cite{Redmon} predicts bounding boxes from a single layer, whereas classification and proposal generation share the same base network. SSD~\cite{Liu2015} enhances this approach by predicting bounding boxes from different layers. Faster R-CNN~\cite{Ren2015} simultaneously predicts object bounds and objectness scores at each position.\\
\textbf{Classical traffic light detection:}
There are various publications on traffic light detection from color~\cite{1336354,Kim2013,Diaz-Cabrera2012,Chiang2011,Gong2010}. Mostly, they use a simple thresholding in different color spaces (RGB, HSV). A popular approach from de Charette \etal~\cite{deCharette2009} uses the white top-hat operator to extract blobs from grayscale images. Other methods use the shape of the traffic light bulb for proposal generation~\cite{1336354, Omachi2009, Omachi2010}. Stereo vision is used as an additional source in~\cite{7995756}. Other publications also use a priori information from maps~\cite{37259, Levinson2011, Barnes2015} in their recognition system. Multi-camera systems are described in~\cite{8317946} and~\cite{7995852}. \\
\textbf{Traffic light detection by CNNs:} As the first method for detecting traffic lights from CNNs the DeepTLR~\cite{7535408} was presented in 2016. Their net returns a pixel-segmented image on which they apply a bounding box regression for each class. Bach \etal~\cite{7995852} also use a CNN segmentation for object detection in a multi-camera setup. Behrendt \etal~\cite{7989163} present a complete detection, tracking and classification system based on CNNs and deep neural networks.   

\section{TL-SSD}
\label{sec:tlssd}
This chapter presents the TL-SSD, a modified single shot detector for traffic light detection. Section~\ref{sec:principle} presents the basic principle of SSD. Section~\ref{sec:base_network} explains why we used Inception  instead of VGG as a base network. Section~\ref{sec:dilemma} discusses and analyzes the problem of SSD for small objects. It shows an adaption to also detect small objects without increasing the input image size. Section~\ref{sec:receptive} analyzes the receptive field of the net and clarifies, in which layers which objects can be detected. Section~\ref{sec:nms} briefly explains how to adapt the non-maximum suppression to avoid multiple detections on a single object. Section~\ref{sec:state} deals with modifications made for state prediction. 
\subsection{Basic Principle}
\label{sec:principle}
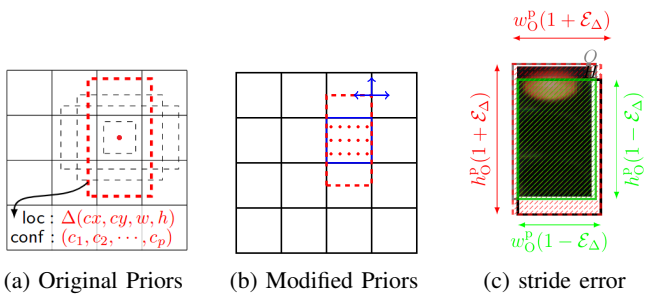
\begin{figure}
    \centering

    \begin{subfigure}[b]{0.29\linewidth}
        \includegraphics[width=\linewidth, trim=0cm 0 10 0cm]{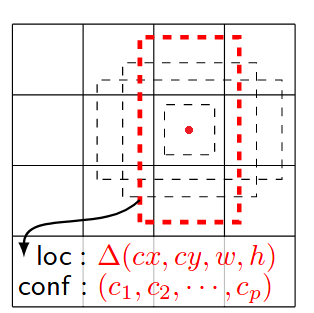}
        \caption{Original Priors}
        \label{fig:priors_before}
    \end{subfigure}
      \hfill
    \begin{subfigure}[b]{0.29\linewidth}
    \scalebox{0.6}{
        \begin{tikzpicture}[yshift=-5cm]
        \draw [line width = 1pt] (0.0,0.0) rectangle (4.0, 4.0);
        \draw [line width = 1pt] (0.0,1) -- (4,1);
		\draw [line width = 1pt] (0.0,2) -- (4,2);
		\draw [line width = 1pt] (0.0,3) -- (4,3);
		\draw [line width = 1pt] (1.0,0) -- (1,4);
		\draw [line width = 1pt] (2.0,0) -- (2,4);
		\draw [line width = 1pt] (3.0,0) -- (3,4);
		
		\draw [line width = 1pt, blue] (2,2) rectangle (3.0, 3.0);
		\fill[red] (2.9, 2.5) circle [red, radius=1pt];
		\fill[red] (2.7, 2.5) circle [red, radius=1pt];
		\fill[red] (2.5, 2.5) circle [red, radius=1pt];
		\fill[red] (2.3, 2.5) circle [red, radius=1pt];
		\fill[red] (2.1, 2.5) circle [red, radius=1pt];
		\fill[red] (2.9, 2.8) circle [red, radius=1pt];
		\fill[red] (2.7, 2.8) circle [red, radius=1pt];
		\fill[red] (2.5, 2.8) circle [red, radius=1pt];
		\fill[red] (2.3, 2.8) circle [red, radius=1pt];
		\fill[red] (2.1, 2.8) circle [red, radius=1pt];
		\fill[red] (2.9, 2.2) circle [red, radius=1pt];
		\fill[red] (2.7, 2.2) circle [red, radius=1pt];
		\fill[red] (2.5, 2.2) circle [red, radius=1pt];
		\fill[red] (2.3, 2.2) circle [red, radius=1pt];
		\fill[red] (2.1, 2.2) circle [red, radius=1pt];
		\draw [line width = 1.5pt, red,dashed] (2,1.5) rectangle (3.0, 3.5);
		\draw [->,line width = 1pt, blue] (3.0,3.5) -- (3,3.9);
		\draw [->,line width = 1pt, blue] (3.0,3.5) -- (3.4,3.5);
		\draw [->,line width = 1pt, blue] (3.0,3.5) -- (2.6,3.5);

                \end{tikzpicture}}
        \caption{Modified Priors}
        \label{fig:priors_after}
    \end{subfigure}
    \hfill
        \begin{subfigure}[b]{0.29\linewidth}
            \includegraphics[width=\linewidth]{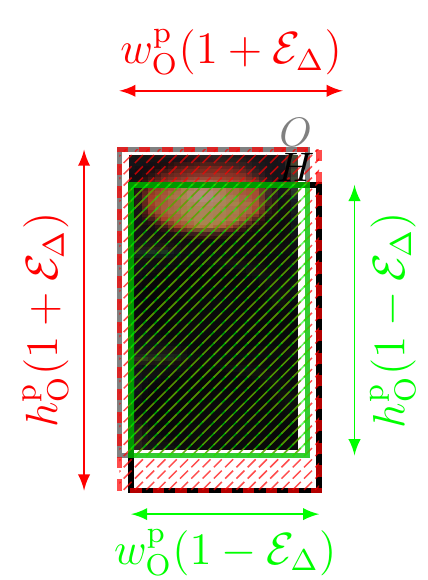}
            \caption{stride error}
            \label{fig:shifting_error}
        \end{subfigure}
    \caption{Comparison of original (a) and modified priors (b). The original priors are placed in the center of each feature cell. In order to also detect smaller objects, the stride has to be smaller. Therefore we adapt the priors for arbitrary positions in the feature cell. }\label{fig:priors}
\end{figure}

Figure~\ref{fig:ssd_net} shows an illustration of the SSD architecture used for prediction of bounding boxes.
Single shot detection is based on a common CNN for feature extraction. In addition, one or several convolutional layers are set up on existing feature maps and predict a fixed number of bounding boxes and a corresponding confidence score. The predicted boxes are offsets with respect to predefined and fixed prior boxes. Prior boxes are distributed over a feature map and located in the center of each cell (see Figure~\ref{fig:priors_before}). 
\\\textbf{Training:} SSD optimizes both, localization and confidence loss $L_{conf}$ as a weighted sum leading to the loss function
\begin{align}
L=\frac{1}{N}(L_{conf} + \alpha \cdot L_{loc})
\label{eq:ssd_loss}
\end{align}
for $N$ matched prior boxes. The localization loss $L_{loc}$ is a regression to offsets between prior boxes and prediction. Position (center) as well as size (width, height) is included. The confidence loss is calculated as a softmax with cross-entropy loss. Details are given in~\cite{Liu2015}.

\subsection{Inception-v3 instead of VGG}
\label{sec:base_network}
Original SSD uses the well-established VGG-16 as the base network. However, traffic light recognition is a task which requires both, high accuracy and real-time capability. As shown in~\cite{DBLP:journals/corr/CanzianiPC16}, there exist networks with a better accuracy vs speed trade-off than VGG. We decided to use Inception-v3~\cite{DBLP:journals/corr/SzegedyVISW15} as one of the base networks with the highest top-1 accuracies at moderate speed. Further benefits are the Inception modules (Figure~\ref{fig:ssd_net}), which concatenate different receptive fields (see Table~\ref{tab:layer_sizes}). Thereby, context information and local information are combined, which is needed to determine the traffic light state (local information) and the existence of a traffic light (context information, see Section~\ref{sec:receptive}).

\subsection{The Dilemma of Small Objects}
\label{sec:step}

The design of convolutional neural networks tends to rising depth with more convolutional layers as later layers are known to generate richer features than early layers~\cite{Szegedy}. During training of SSD, predefined prior boxes are matched with the ground truth objects. Whereas the size(s) and aspect ratio(s) can be chosen arbitrarily, the stride of each default box is defined by the size of the feature layer. Original SSD places the center of a prior box at $\underline{c}_p = \Big(\frac{x+0.5}{w_{\mathrm{f}}}, \frac{y+0.5}{h_{\mathrm{f}}} \Big)$,
where $w_{\mathrm{f}}\cdot h_{\mathrm{f}}$ is the size of the chosen feature layer and $x$ and $y$ are the feature layer coordinates, respectively.
 Table~\ref{tab:layer_sizes} illustrates the sizes as well as the corresponding prior box stride of important convolutional layers of the Inception\_v3 net, which we use for our main experiments. In late inception layers, strides of 8, 16 or 32 pixels with regard to the input image are set. Those high strides lead to a high risk of missing traffic light objects. We analyze the maximum allowed step with respect to the smallest to be detected object in the following.  \\
 \textbf{Stride with respect to object size:} Figure~\ref{fig:shifting_error} illustrates a traffic light ground truth circumscribed by an ideal bounding box O. The black hypothesis H shows a not perfectly aligned detection due to a higher bounding box stride. We denote the positioning error as $\mathcal{E}_\Delta$. To express the effect of a too high step size on the detection accuracy, we use the metric intersection over union defined as 
 
  \begin{equation}
  \theta_{\mathrm{IOU}}= \frac{|H \cap O|}{|H \cup O|} = \frac{|H \cap O|}{|H| + |O| - |H \cap O|}.  
  \label{iou_theta}
  \end{equation} 
  It expresses the overlap between a ground truth and detection bounding box. A common threshold counting an object as detected is $\theta_{\mathrm{IOU}}=0.5$. Our goal is to determine the allowed stride $\Delta=2\mathcal{E}_\Delta$ with respect to the IoU we want to reach. The IoU with respect to the stride error can be derived according to Figure~\ref{fig:shifting_error} as 
  
  \begin{align}
  \theta_{\mathrm{IOU}}=\frac{(1-\mathcal{E}_\Delta)^2}{(1+\mathcal{E}_\Delta)^2}
  \end{align}
\begin{table}[!t]
\centering
\caption{Import layer sizes of Inception\_v3 and the corresponding feature map size. After the two inception c modules, the original image sized is reduced by a factor of 32.  }
\label{tab:layer_sizes}
\scalebox{0.85}{
\begin{tabular}{lccccc}
\toprule
\textbf{layer} & \textbf{height} & \textbf{width} & \textbf{ratio} & \textbf{stride} & \textbf{receptive field (min/max)} \\ 
\midrule 
\textbf{Input} & 512 & 2048 & 1 & 1 & 1x1  \\ 
\textbf{conv\_1} & 255 & 1023 &$ \sim 1/2$ & 2 & 3x3\\ 
\textbf{conv\_2} & 253 & 1021 & $\sim 1/2$ & 2 & 7x7\\ 
\textbf{conv\_3} & 253 & 1021 & $\sim 1/2$ & 2 & 11x11\\ 
\textbf{conv\_4} & 124 & 508 & $\sim 1/4$& 4 & 23x23\\
\midrule 
\textbf{inception\_a1} & 62 & 254 & $\sim 1/8$ & 8 & 31x31 / 63x63\\ 
\textbf{inception\_a2} & 62 & 254 & $\sim 1/8$ & 8 & 31x31 / 95x95\\ 
\textbf{inception\_a3} & 62 & 254 & $\sim 1/8$ & 8 & 31x31 / 127x127\\
\midrule
\textbf{inception\_b1} & 31 & 127 & $\sim 1/16$ & 16 & 47x47 / 351x351\\ 
\textbf{inception\_b2} & 31 & 127 & $\sim 1/16$ & 16 & 47x47 / 543x543\\ 
\textbf{inception\_b3} & 31 & 127 & $\sim 1/16$ & 16 & 47x47 / 735x735\\ 
\textbf{inception\_b4} & 31 & 127 & $\sim 1/16$ & 16 & 47x47 / 927x927\\ 
\midrule
\textbf{inception\_c1} & 15 & 63 & $\sim 1/32$ & 32 & 79x79 / 1183x1183\\ 
\textbf{inception\_c2} & 15 & 63 & $\sim 1/32$ & 32 & 79x79 / 1311x1311\\ 
\bottomrule
\end{tabular}}
\end{table}
 leading to
  \begin{align}
  \label{eq:delta}
  \mathcal{E}_\Delta  = \frac{\theta_{\mathrm{IOU}}- 2 \sqrt{\theta_{\mathrm{IOU}}}+1}{1-\theta_{\mathrm{IOU}}}.
  \end{align}
In order to reach $\theta_{\mathrm{IOU}}=0.5$ a step size of $\Delta=0.34$ is necessary. In consequence, a maximum stride of $0.34\cdot 5 ~\text{pixels} = 1.7 ~\text{pixels}$ is needed to guarantee a detection of objects with a width of 5 pixels. As seen in Table~\ref{tab:layer_sizes}, only layer conv\_1 - conv\_3 can satisfy this condition. From experience, those layers do not provide strong enough features for accurate detection and a small number of false positives.\\
\textbf{Priorbox stride adaption:} Therefore, we propose to adapt the prior box centers. We create an arbitrary number of priors per feature cell which can be described by   
\begin{align}
\label{eq:offset}
\underline{c}_{\mathrm{p}} = \Big(\frac{x+\underline{o}_x}{w_{\mathrm{f}}}, \frac{y+\underline{o}_y}{h_{\mathrm{f}}} \Big), \quad  0 \leq o_{x,i}, o_{y,i} \leq 1,
\end{align}
where $\underline{o}_x$ and $\underline{o}_y$ are offset vectors in the feature cell domain. We propose to define $o_{y,i}=r \cdot o_{x,i}$ due to the results obtained from Equation\,(\ref{eq:delta}). Figure~\ref{fig:priors_after} illustrates examples for possible priors (red) in one single cell (blue) with an aspect ratio of 3. By means of this adaption, the stride is independent of the feature layer size.
\label{sec:dilemma}

\begin{figure}[h]
    \centering

	\begin{subfigure}[b]{0.24\linewidth}
	\includegraphics[width=\linewidth,height=\linewidth, trim=0cm 0 10 0cm]{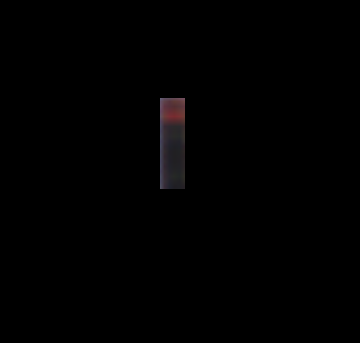}

	\end{subfigure}
	\hfill
	\begin{subfigure}[b]{0.24\linewidth}
	\includegraphics[width=\linewidth,height=\linewidth]{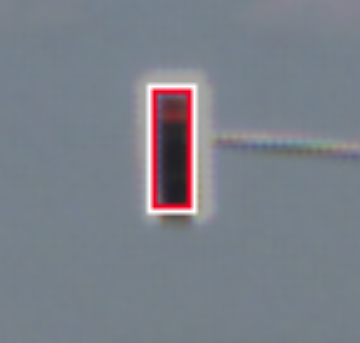}

	\end{subfigure}     
	\begin{subfigure}[b]{0.24\linewidth}
	\includegraphics[width=\linewidth,,height=\linewidth, trim=0cm 0 10 0cm]{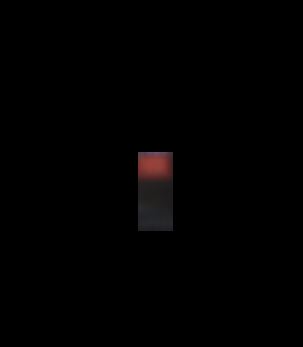}

	\end{subfigure}
	\hfill
	\begin{subfigure}[b]{0.24\linewidth}
	\includegraphics[width=\linewidth,height=\linewidth]{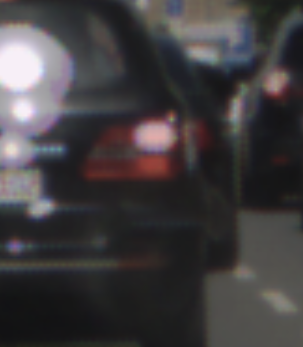}

	\end{subfigure}

    \caption{Traffic light detection needs context information. Especially small objects are not distinguishable from false positives, such as rear lights of vehicles.}
    \label{fig:context}
\end{figure}
\subsection{Receptive Field: How Much Context is Needed?}

\begin{figure}[!t]
\centering
\scalebox{0.7}{
\begin{tikzpicture}
\node[inner sep=0pt] (russell) at (0,0)
    {\includegraphics[width=\linewidth]{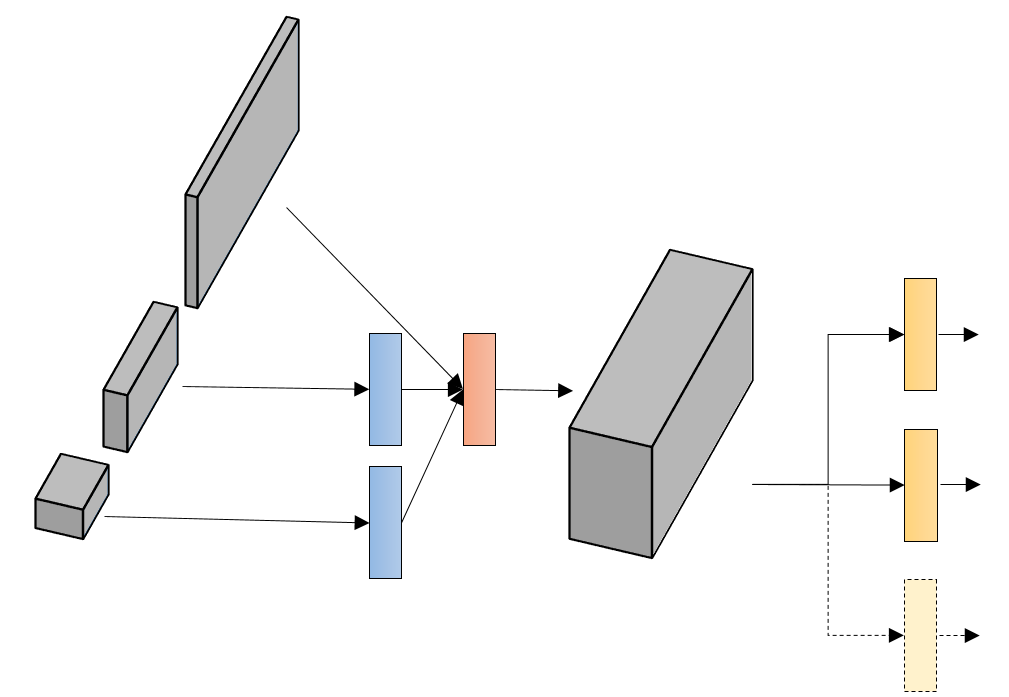}};
    \node[] at (-5,-1.5) {\scriptsize inception\_c2};
        \node[] at (-4.5,-0.5) {\scriptsize inception\_b4};
                \node[] at (-3.5,1.5) {\scriptsize inception\_a3};
    
\node[rotate=90, xshift = 0.3cm, yshift = -0.15cm] at (-1.15,-1.8) {\scriptsize interp};
\node[rotate=90, xshift = 0.3cm, yshift = -0.15cm] at (-1.15,-0.7) {\scriptsize interp};
\node[rotate=90, xshift = 0.3cm, yshift = -0.19cm] at (-0.42,-0.7) {\scriptsize concat};
\node[rotate=90, xshift = 0.4cm, yshift = -0.38cm] at (3.17,-0.3) {\scriptsize conv};
\node[rotate=90, xshift = 0.4cm, yshift = -0.4cm] at (3.17,-1.6) {\scriptsize conv};
\node[rotate=90, xshift = 0.4cm, yshift = -0.4cm] at (3.17,-2.85) {\scriptsize conv};
\node[yshift = 0.35cm, xshift = 0.4cm] at (4,-1.5) {\scriptsize conf};
\node[yshift = 0.35cm, xshift = 0.4cm] at (4,-0.25) {\scriptsize bbox};
\node[yshift = 0.35cm, xshift = 0.4cm] at (4,-2.8) {\scriptsize state};

\end{tikzpicture}}
\caption{Feature layer concatenation for fusion of local and context information. We concatenate early and late layers and use them for bounding box and confidence prediction. Additional state prediction can be performed by another convolutional layer.  }
\label{fig:concat}
\end{figure}
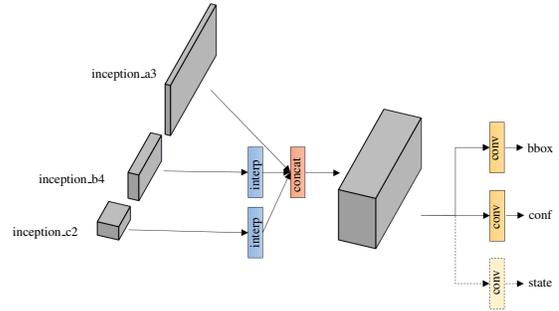

The receptive field of CNNs can be explained as the region in the input image a feature is "looking at". Increasing the receptive field is done by either applying kernels in a convolutional layers or using pooling layers.\\
The core idea behind SSD is to detect larger objects in late layers as late layers "look" at larger regions in the input image. Nevertheless, besides the size of the object, different object types require more context information than others.\\
Traffic lights have a visual appearance, which seems to be unique and easy to detect at first glance. However, Figure~\ref{fig:context} illustrates that traffic lights are hard to differentiate from background without any context information. Especially at rear lights of vehicles, window panes or in trees many potential false positives appear. Adding context information as in Figure~\ref{fig:context} makes better delimitation possible. For a better understanding the receptive field of important layers is given in Table~\ref{tab:layer_sizes}. Please note, that we analyzed the theoretical instead of effective receptive field, see~\cite{Luo:2016:UER:3157382.3157645}. The receptive field for each layer is calculated as
\begin{align}
r_f = r_{\mathrm{f,in}} + (k-1)s_{\mathrm{in}},
\end{align}
where $r_{\mathrm{f,in}}$ is the receptive field of the previous layer, $k$ is the kernel size and $s_{\mathrm{in}}$ is the stride, i.e. ratio between feature layer size and the input image.\\
\textbf{Feature layer concatenation:} Adding context helps to reduce false positives but can also lead to a loss of an accurate position of the bounding box location. Furthermore, state determination (color of the lamp) is an information, for which clearly less context is required. In order to obtain all relevant information, we propose to use feature layer concatenation for bounding box prediction as shown in Figure~\ref{fig:concat}. PSPNet~\cite{Zhao2016} has shown the effectiveness of feature layer concatenation for semantic segmentation.  Therefore we combine early and late feature layers so that early layers can provide accurate location and state information and late layers make the decision if traffic light or not. Therefore, the late layers are interpolated. Box and confidence prediction is done similar to original SSD.

\label{sec:receptive}
\subsection{Non-Maximum Suppression}
\label{sec:nms}
In the original SSD, a non-maximum suppression (NMS) is used to suppress multiple detections on a single object. We decode the state of the traffic light as a separate class. Doing the NMS class-wise leads to multiple detections (same position but differing in state and confidence) of one traffic light instance. We adapt the NMS to a class-independent one. The final state of the traffic light is picked from RoI number
\begin{align}
s_i = \operatorname*{arg\,max}_{c_i~\in ~\mathbf{C}} c_i,
\end{align} 
where $\mathbf{C}$ are all elements assigned to one real object. In other words, we pick the class of the RoI with the highest confidence value. All elements in $\mathbf{C}$ meet the following constraint $IoU(\mathbf{C},b_i) \geq 0.35,$
i.e. if the overlap of a prediction is larger than a 0.35. We decided for a relatively small overlap threshold as we have a high number of small objects in the dataset and high overlaps are hard to reach for small objects.  
\begin{figure*}[!t]
    \centering

    \begin{subfigure}[b]{0.32\textwidth}
        \includegraphics[width=\textwidth, trim=0cm 0 10 0cm]{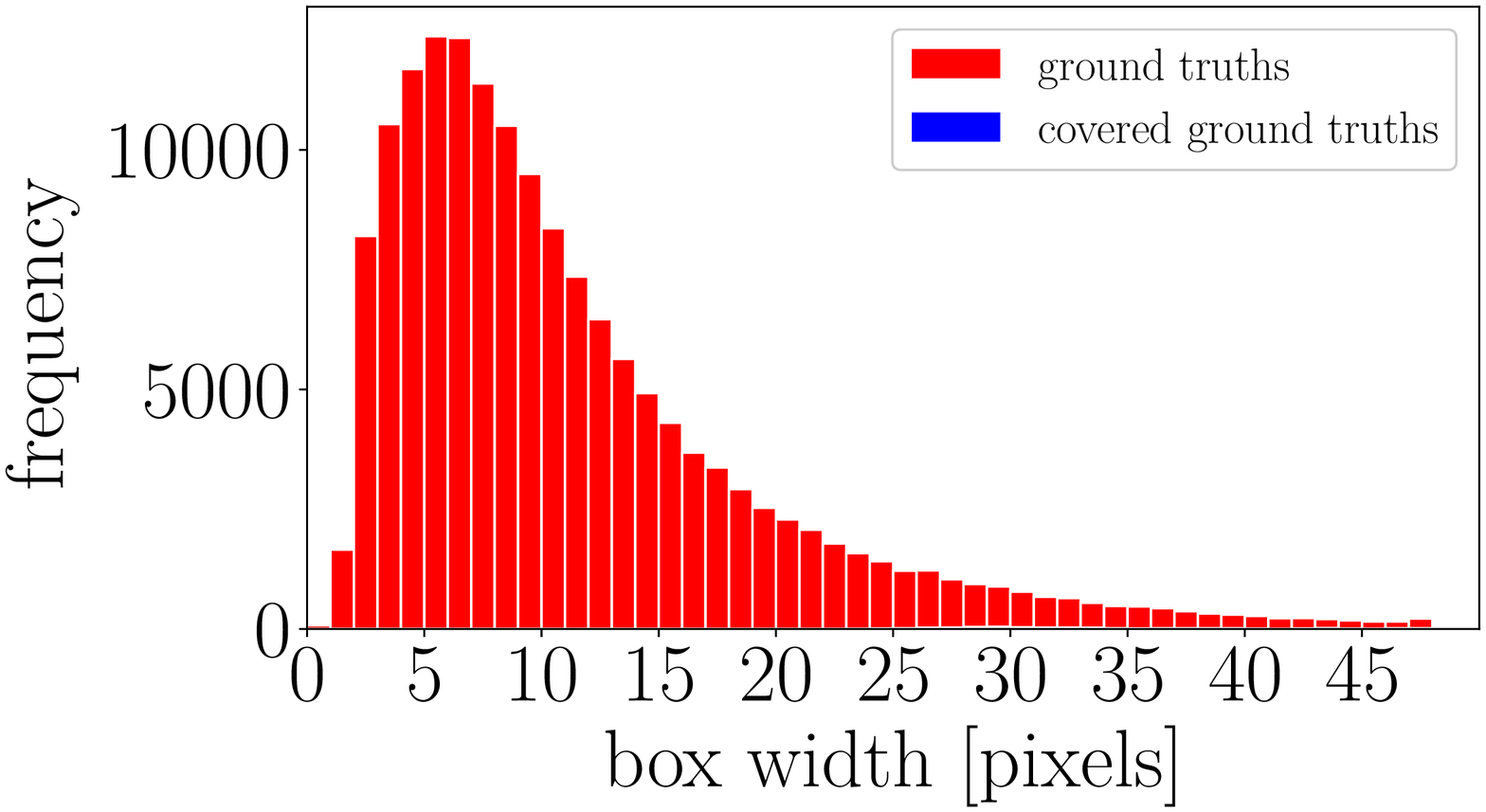}
        \caption{original (VGG)}

    \end{subfigure}
      \hfill
          \begin{subfigure}[b]{0.32\textwidth}
              \includegraphics[width=\linewidth, trim=0cm 0 10 0cm]{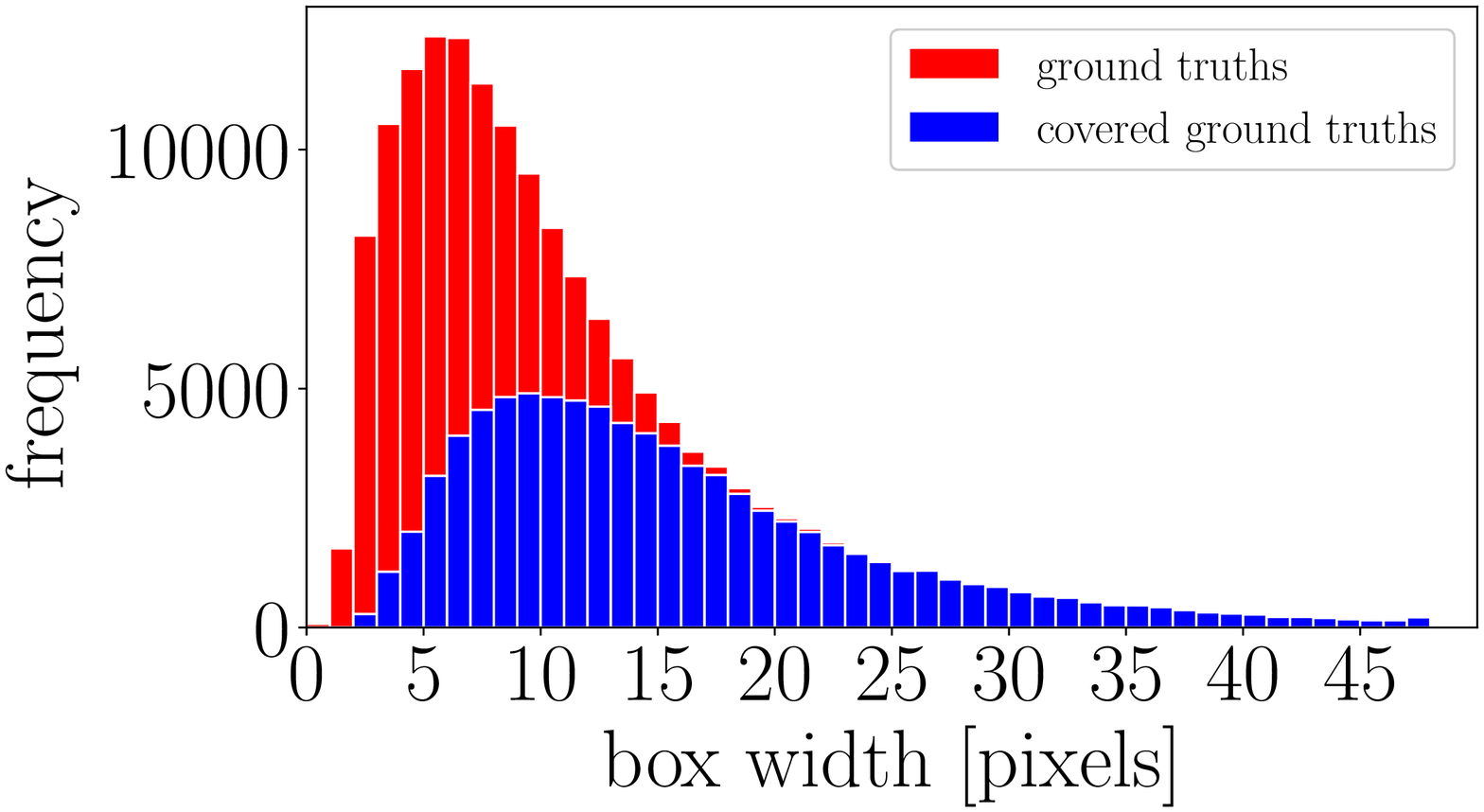}
              \caption{Adapted sizes and ratios}

          \end{subfigure}
    \hfill
            \begin{subfigure}[b]{0.32\textwidth}
                \includegraphics[width=\linewidth, trim=0cm 0 10 0cm]{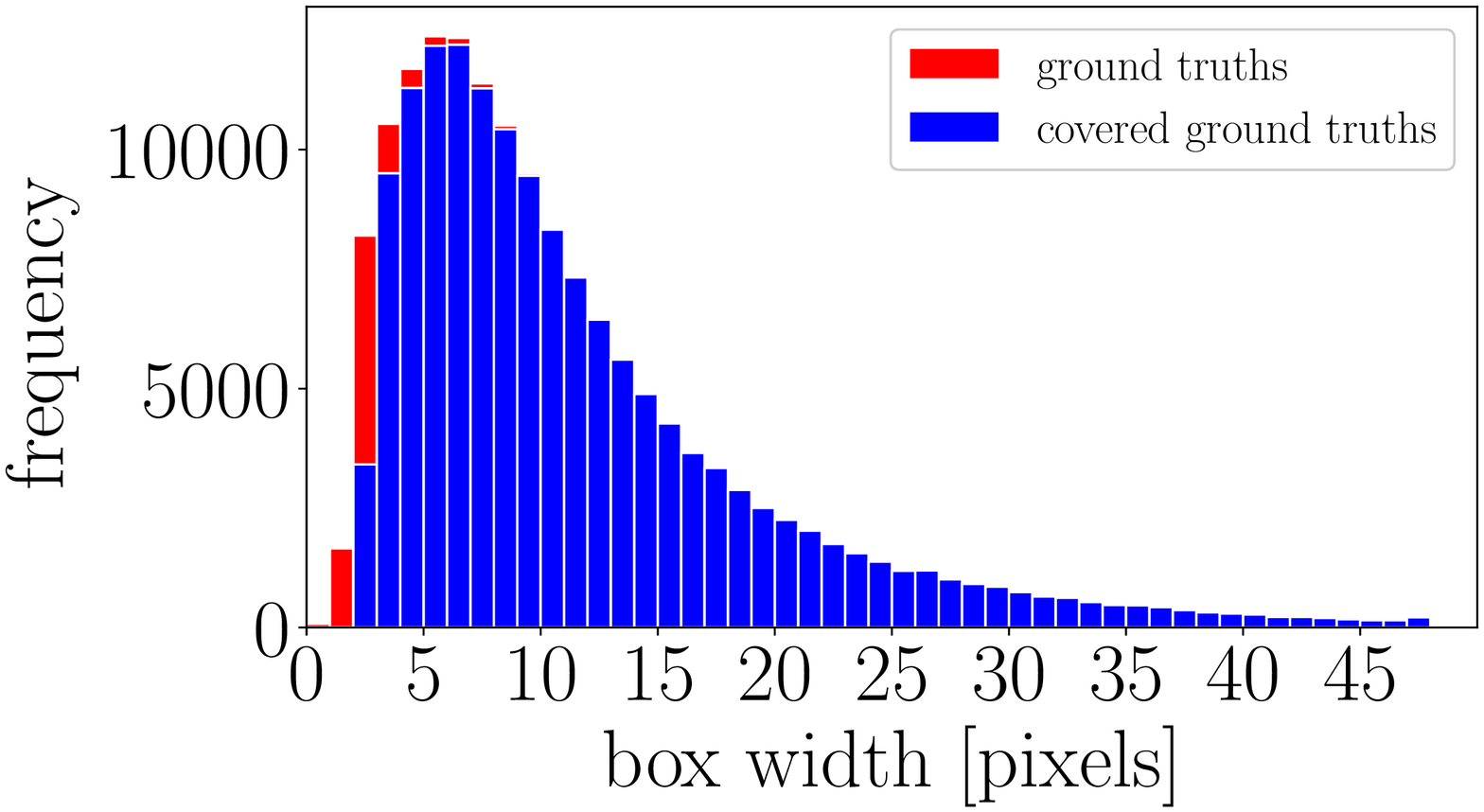}
                \caption{Adapted stride, size and ratios}

            \end{subfigure}
    \caption{Matched ground truths with respect to the bounding box width. The left figure shows the poor coverage of the original SSD prior boxes applied on the DriveU Traffic Light Dataset. After adapting the sizes and aspect ratios, a much better coverage is reached. However small objects can not be covered due to the high stride (16, see Table~\ref{tab:layer_sizes}). After adaption of the prior box layer, we reach the coverage in (c), in which objects down to a width of 3 pixels are mostly covered.}
    \label{fig:coverage}
\end{figure*}

\subsection{Extensions for State Prediction}
\label{sec:state}
We tried two different ways to additionally predict the state of traffic lights. A first approach was to replace the binary classification task by a multi-class task, in which we assign each label to one state class, i.e. $c_i=(\mathrm{off, red, yellow, green})$. In total SSD predicts 5 classes by adding one additional background class. The network was trained similar to the binary classification task according to Equation\,(\ref{eq:ssd_loss}). Although this enabled a state detection, we recognized a significant decrease in accuracy in the pure detection task (traffic light vs background). One possible explanation is that the foreground confidence is distributed over all states which leads to a more dominant background confidence.
In oder to avoid this problem we adapted the SSD approach as follows: The confidence layer still performs the binary classification (traffic light vs background), whereas according to Figure~\ref{fig:concat} an additional layer performing state prediction is added. Optimization is done by using a separate state loss $L_{\mathrm{state}}$ leading to the overall loss given as    
\begin{align}
L=\frac{1}{N}(L_{\mathrm{conf}} + \alpha \cdot L_{\mathrm{loc}} + \beta \cdot L_{\mathrm{state}}).
\end{align} with a state weight factor $\beta$.
Unlike the confidence loss, which is calculated as a softmax with cross-entropy loss, we use a sigmoid loss function defined as
\begin{align}
L_{\mathrm{state}} = \frac{1}{1 + e^{-s_i}}
\end{align}
over multiple state confidences $s_i$. ~\cite{DBLP:journals/corr/abs-1804-02767} have shown that this can be beneficial for highly correlated classes (such as the gender of humans or the state of traffic lights). This way, the pure detection accuracy is not affected compared to binary training and an additional state prediction is possible.
\section{Experiments}
\label{sec:experiments}
\subsection{Dataset}
\label{sec:dataset}

For training of our TL-SSD approach we use the DriveU Traffic Light Dataset~\cite{DBLP:conf/icra/Fregin}. This dataset is the largest published containing  more than 230\,000 hand-labeled traffic lights.\\
\textbf{Training set:} The dataset includes over 300 different classes with very specific tags such as relevancy, color, number of lamps, orientation and pictogram (pedestrian, tram, arrows ...). For practical usage of a traffic light detection system, mainly front orientated traffic lights, i.e. traffic lights facing the vehicles road have to be detected. Traffic lights for pedestrians, trams or turned traffic lights (valid for the oncoming traffic) are negligible. During training, we calculate the confidence loss $L_{\mathrm{conf}}$ as a two-class problem (traffic light vs. background) and the state loss $L_{\mathrm{state}}$ as a 4-class problem (red, yellow, green, off).
\begin{table}[!b]
\centering
\caption{DTLD statistics used for training and evaluation.}
\label{tab:dataset}
\begin{tabular}{lcccc}
\toprule
&\textbf{Cities} & \textbf{Images} & \textbf{Sequences} & \textbf{Objects} \\
\midrule 
\textbf{Training} & 11 & 28526 & 1478 & 159902 \\
\textbf{Evaluation} & 11 & 12453 & 632 & 72137 \\
\bottomrule
\end{tabular}
\end{table}
\\ \textbf{Evaluation set:} For evaluation we use the proposed split of the dataset. Thus, the evaluation set contains around one third of all annotations. Detailed statistics of both sets can be seen in Table~\ref{tab:dataset}. The term sequences describes one unique intersection with a varying number of unique traffic lights.\\
\textbf{Limitations of the Evaluation Set:} Varying label rules is one key problem of the dataset for the purpose of evaluation. The majority of the images are annotated with front-facing traffic lights only. A small part also contains annotated turned traffic lights (e.g. for pedestrians or oncoming traffic). Detections on those traffic lights are counted as false positives as they are not consistently annotated. Nevertheless, it is the largest and most carefully annotated dataset predestined for our use-case.\\
\textbf{Don't-care (dc) objects:} For evaluation we apply several filters. All traffic lights not tagged as \emph{front}, i.e. traffic lights not valid for the direction of the vehicle, are set as dc objects. Traffic lights valid for pedestrians, cyclists, trams or buses are also tagged as dc. In some experiments we use a minimum detection width and also tag all smaller annotations as dc. Detections on dc objects are not counted as false positives, but not detected dc objects are also not counted as false negatives.

\subsection{Metrics}

\textbf{Recall/True positive rate:} We express the percentage number of detected traffic lights by the \emph{true positive rate}, or also called \emph{recall} defined as $P_{TP} = \frac{\mathrm{TP}}{\mathrm{TP}+\mathrm{FN}}$, where a true positive (TP) is counted for an overlap threshold IoU larger than 0.3 or 0.5 according to Formula (\ref{iou_theta}). False positives (FP) are counted for predictions not overlapping with a ground truth by the defined overlap threshold. Multiple detections on one ground truth object are also counted as false positives. We evaluate all trainings by an ROC curve (miss rate $P_{FN}$ vs. false positives per image (FPPI)). The running parameter of the ROC curve is the confidence threshold $c_i$, which is between 0 and 1. We evaluate for different IoU threshold values. Thus, the ROC curve can be written as $P_{FN}(FPPI, c_i, IoU)$. \\
\textbf{Log-average miss rate:} In order to compare trainings by one single metric, we use the \emph{log-average miss rate} 
\begin{align}
LAMR = \frac{P_{FN}(10^{-1}) + P_{FN}(10^{0}) + P_{FN}(10^{1}) }{3},
\end{align}
where $P_{FN}=1-P_{TP}$ and three characteristic points of the ROC curve are picked. It is a metric also used in many popular pedestrian detection publications (see~\cite{5206631},\cite{5975165}). We picked FPPI values of $10^{-1},~10^0,~10^1$ as they correspond to suitable operating points for subsequent modules (e.g. tracking). 
\subsection{Coverage between Prior Boxes and Ground Truths}
SSD only trains ground truths covered by at least one prior box with a minimum intersection over union value. We choose an IoU threshold of 0.3 to also cover very small annotations. Figure~\ref{fig:coverage}(a) illustrates the poor coverage without any adaptations using the original SSD parameters. Figure (b) shows improved coverage when adapting the size and aspect ratio of the prior boxes. We picked a fixed aspect ratio of 0.3 and multiple widths from 4 up to 38 pixels. However, as prior boxes are generated in layer inception\_b4, the stride (16, see Table~\ref{tab:layer_sizes}) is too high to also cover small objects. With our adaptations described in Section~\ref{sec:step} the final coverage of Figure~\ref{fig:coverage}(c) can be reached. We choose the offset vectors (Equation\,(\ref{eq:offset})) according to the derived allowed step size of Equation\,(\ref{eq:delta}) leading to offsets of 0.16 in this layer, which corresponds to 2-3 pixels in width and 6-9 pixels in height.  

\subsection{Deeper Network - Better results?}
\begin{figure}[!t]
\centering
\includegraphics[width=0.86\linewidth]{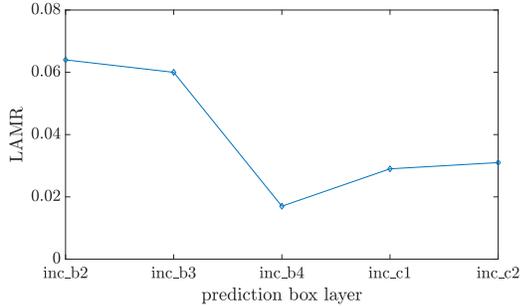}
\caption{Influence of the network depth on the LAMR result. Predicting boxes in Inception\_b4 yields the best results. }
\label{fig:depth_lamr}
\end{figure}

For this experiment, we trained networks with one prior box and prediction layer only creating all desired sizes. 
In order to guarantee a fair comparison, we adapted the step offset parameters (half step offset in b-layers because of half stride compared to c-layers). Figure~\ref{fig:depth_lamr} illustrates the log-average miss rate for different prior box depth. As expected, too early layers have weak features and less context. Late layers have a too large context leading to a loss in detailed information. The sweet spot is layer inception\_b4 with a LAMR of 0.02.


\subsection{Using multiple output layers}
Original SSD achieved better results by using multiple layers for bounding box prediction. We did several experiments on using multiple layers for detection, in which later layers have to detect larger traffic lights and smaller layers have to detect smaller objects. Results can be seen in Table~\ref{tab:results}. The results do not correspond to the findings of the SSD authors. The best results are comparable to the best results of one layer only. One possible explanation is that all objects are in a comparable size range. Furthermore, requirements on the receptive field are approximately equal as small objects typically need a higher receptive field than large objects.
\subsection{Does more data help?}
\begin{figure}[!t]
\centering
\includegraphics[width=0.86\linewidth]{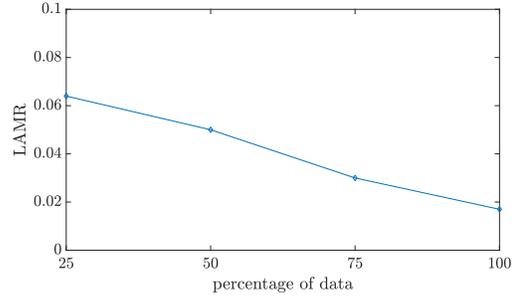}
\caption{Influence of the amount of data on the LAMR.}
\label{fig:data_lamr}
\end{figure}

One common statement about CNNs is that more data automatically helps to improve generalization and overall results. To investigate the impact of the amount of data on the recall, we take the best model of the previous result. We generate prior boxes in layer inception\_b4. We generate four sub-training sets consisting of 25, 50, 75 and 100 \% of the training set. Figure~\ref{fig:data_lamr} illustrates the LAMR result for all four subsets. An almost linear relation can be seen, which clarifies than even more data would enhance the detection results. The point of saturation is not yet reached.
\setlength{\tabcolsep}{2pt}
\begin{table}[!b]
\centering
\caption{All results of TL-SSD. Check marks illustrate on top of which layer(s) predictions are made. Our prior box stride adaptations increased LAMR by 4 percent in layer Inception\_b4. }
\label{tab:results}
\scalebox{0.7}{
\begin{tabular}{cccccc|c|c}
\cline{1-3}
\hline
 \multicolumn{6}{c|}{\textbf{Prior Box Layer}} &  \textbf{LAMR (IoU=0.5)}& \textbf{Runtime [ms]}\\
 \hline
\rot{Inception\_b1} & \rot{Inception\_b2} & \rot{Inception\_b3} & \rot{Inception\_b4} & \rot{Inception\_c1} & \rot{Inception\_c2} &  \\
\hline
&\cmark & & & & & 0.064  & 101 \\
& & \cmark  & & & & 0.060  & 106\\

& & & \cmark  & & & \textbf{0.017}  & 111\\

& & & ~~$\text{\cmark}^{\Delta}$  & & & 0.046& 105\\

& & & & \cmark  & & 0.029  & 133\\

& & & & & \cmark  & 0.031  & 145\\

\hline
& & &\cmark &\cmark & \cmark  & 0.016 & 121 \\

& & \cmark &\cmark & \cmark  & & \textbf{0.015}  & 117\\
  & \cmark &\cmark & \cmark & & & 0.020  & 122 \\
\hline
 $\text{\cmark}^*$ & & &$\text{\cmark}^*$& $\text{\cmark}^*$& & 0.016  & 165 \\
  ´ & &$\text{\cmark}^*$ &$\text{\cmark}^*$& & & \textbf{0.015}  & 119\\
\hline
\end{tabular}}

\hspace*{-5cm}* concatenation of layers\\
\hspace*{-5cm}$\Delta$ without stride adaption
\end{table} 
\setlength{\tabcolsep}{6pt}  
\subsection{Results over ground truth width}
\begin{figure*}[!t]
    \centering
    \begin{subfigure}[b]{0.95\columnwidth}
        \centering
        \includegraphics[width = 0.96\columnwidth]{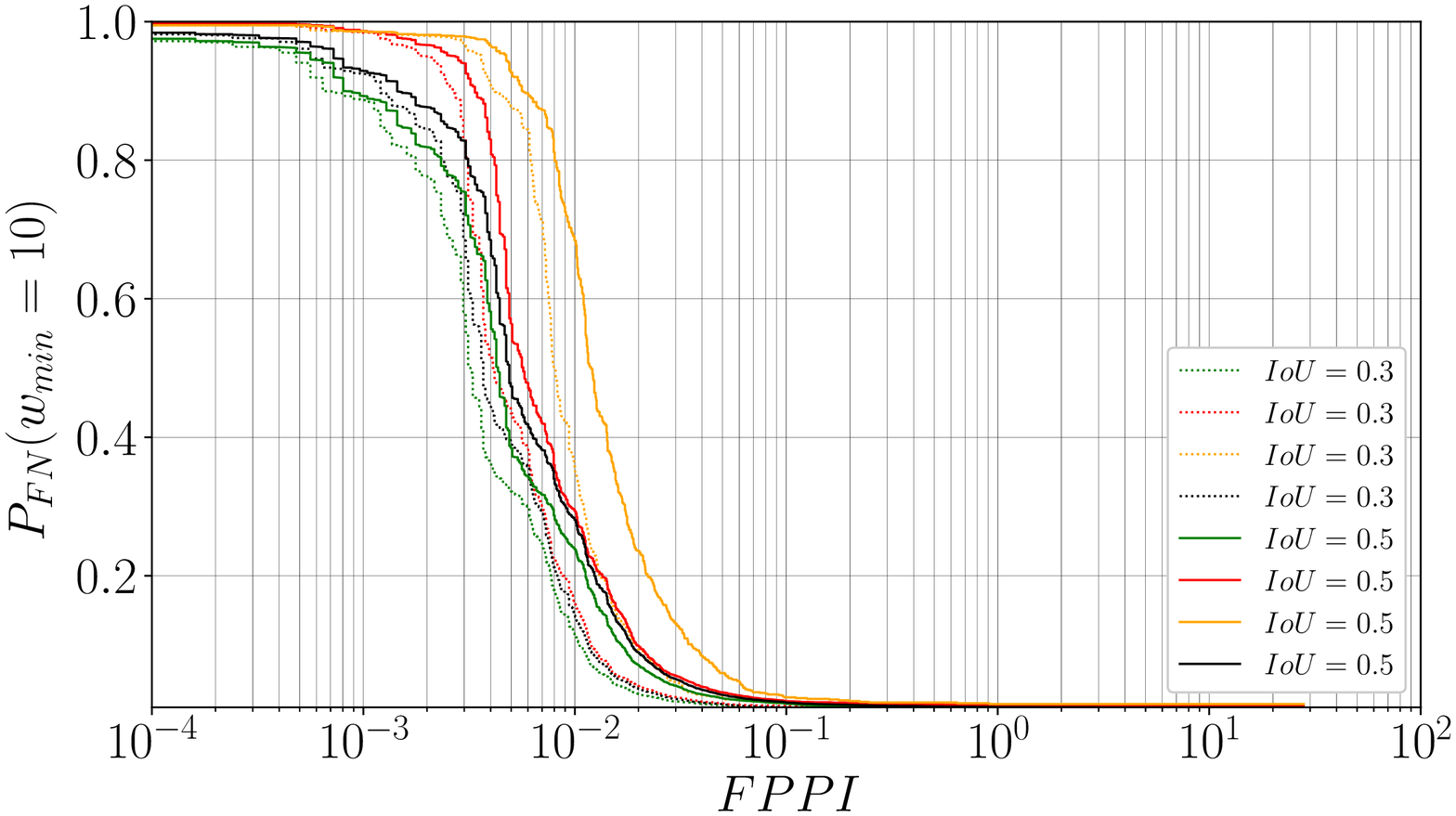} 
        \caption{ROC curves for the pure detection result}

    \end{subfigure}
    \hfill
    \begin{subfigure}[b]{0.95\columnwidth}  
        \centering 
        \includegraphics[width = 0.96\columnwidth]{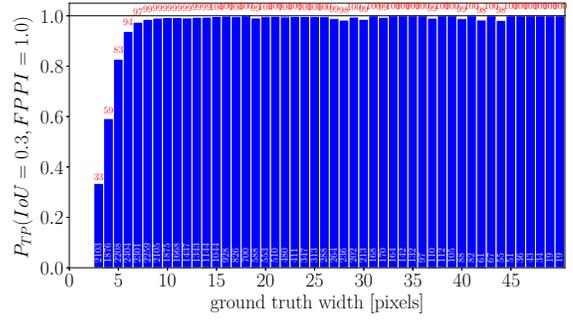}
 		\caption{Detailed detection result for $\mathrm{FPPI}=10^0$}
		\label{fig:res_gt_width}
    \end{subfigure}
 
    \caption{The left figure shows ROC curves for the traffic light states green, yellow, red and all. The discriminant threshold is the confidence value for each prediction, which is between 0 and 1.0. Green traffic lights show a lower miss rate than red traffic lights. Results for a required overlap of 0.3 are clearly better than 0.5, as high overlaps are hardly reachable for small objects. The right figure shows detailed results with respect to the ground truth width. Results are plotted for IoU=0.3 and FPPI=1.0. Recall increases with width. Red numbers show the exact recall value, white values show the absolute number of annotations at the specific width. }
    \label{fig:roc}
\end{figure*}

The detection results with respect to the distance of an object is of particular interest for autonomous driving. In case of traffic lights, a comfortable braking is desired, which requires a detection at high distances. Figure~\ref{fig:res_gt_width} illustrates the detection results with respect to the ground truth width. The respective detection rate is written in red, whereas the absolute number of ground truths at the respective width is written in white. With rising FPPI, the recall increases especially for very small objects. A saturation occurs from approximately ten pixels in width with high recall values.
\subsection{Track-wise Evaluation}

\begin{figure}[!t]
\centering
\includegraphics[width=0.97\linewidth]{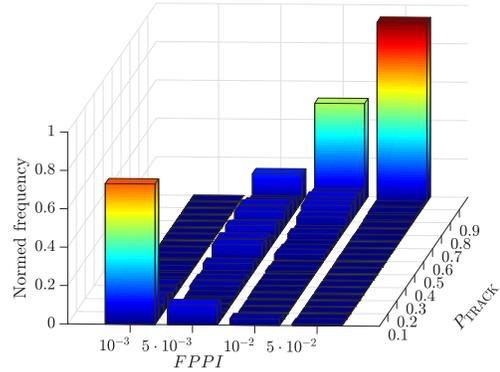}
\caption{Evaluation of the percentage track detection rate. For higher false positive rates, around 95 percent of the objects are detected in 90-100 percent of the frames they appear.}
\label{fig:track_recall}
\end{figure}

DTLD additionally contains track identities, which group unique traffic light instances over multiple frames. A track wise evaluation is of particular interest for a potential system as tracking can compensate missing detections and thus the percentage of detected track is an interesting result. Figure~\ref{fig:track_recall} shows $P_{\mathrm{TRACK}}$ calculated as the number of correct detection of one single track divided by its number of occurrences (i.e. number of frames, in which the object appears). Results show, that for higher FPPI values $P_{\mathrm{TRACK}}=0.9..1.0$ increases up to 95 percent. In other words, 95 percent of all objects in DTLD are detected in almost all frames they occur. There still exist several tracks, which are detected not even once. However, a closer look  shows that, those cases are often tracks only appearing in one single frame or very rare cases (like mirror images of traffic lights).


\section{Conclusion}
\label{sec:conclusion}
In this paper we presented the TL-SSD, an adaption of a single shot detector for traffic light detection and small object detection in general. We adapted the base network from VGG-16 to a faster and more accurate Inception\_v3. Furthermore, we adapted the prior box generation to allow smaller stride in late network layers, which is essential for small object detection. We proved this in a theoretical manner. An adaption of the non-maximum suppression helps to avoid multiple detections on a single object. Furthermore we predict the state of the traffic lights using an additional branch.
Extensive experiments on the DriveU Traffic Light Dataset were presented, which analyzed the properties of our method. We evaluated different operating points differing in the number of false positives per image. We showed, that more data leads to better results and the network depth has to be chosen carefully. Recall values up to 95 percent even for small objects were reached, values increase up to 98-100 percent for larger objects at false positve rates between 0.1 and 10 FPPI.

%

\begin{figure*}[!t]
\centering
	\begin{subfigure}[b]{0.495\textwidth}
	\adjincludegraphics[width=\textwidth,Clip={.00\width} {0.0\height} {0.0\width} {.0\height}]{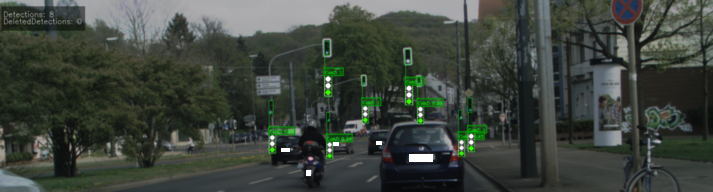}
	    \label{fig:res1}
	\end{subfigure}
	\hfill
	\begin{subfigure}[b]{0.495\textwidth}
	  \adjincludegraphics[width=\textwidth,Clip={.00\width} {0.0\height} {0.0\width} {.0\height}]{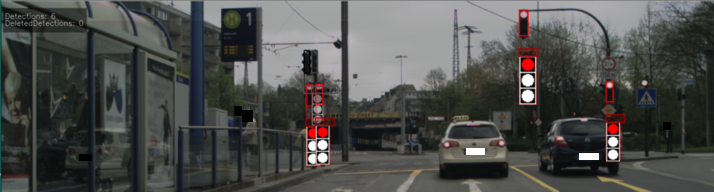}
	  \label{fig:res2}
	\end{subfigure}
	\begin{subfigure}[b]{0.495\textwidth}
	    \adjincludegraphics[width=\textwidth,Clip={.00\width} {0.0\height} {0.0\width} {.0\height}]{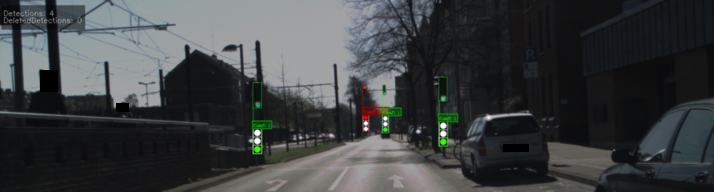}
	    \label{fig:res3}
	\end{subfigure}
	\hfill
	\begin{subfigure}[b]{0.495\textwidth}
	\adjincludegraphics[width=\textwidth,Clip={.00\width} {0.0\height} {0.0\width} {.0\height}]{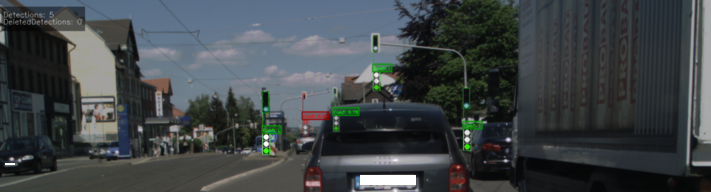}
	\label{fig:res4}
	\end{subfigure}

	\begin{subfigure}[b]{0.495\textwidth}
		\adjincludegraphics[width=\textwidth,Clip={.00\width} {0.0\height} {0.0\width} {.0\height}]{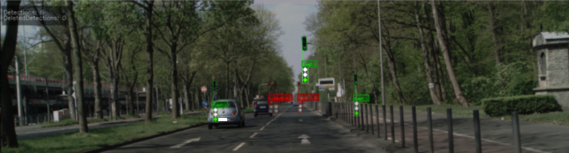}
		\label{fig:res4}
	\end{subfigure}
	\hfill
	\begin{subfigure}[b]{0.495\textwidth}
	 \adjincludegraphics[width=\textwidth,Clip={.00\width} {0.0\height} {0.0\width} {.0\height}]{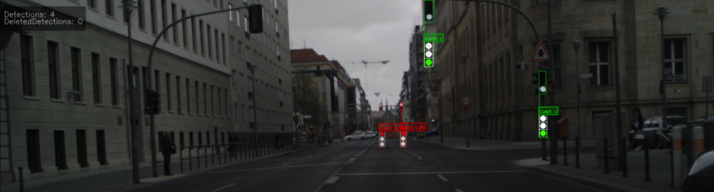}
	  \label{fig:prediction6}
	\end{subfigure}
	\caption{Prediction results from our trained SSD on the DriveU Traffic Light Dataset. The color of the bounding box indicates the predicted state. Very small objects with a few pixels in width can be detected after all adaptations described in this paper. }
	\label{fig:prediction}
\end{figure*}

%
%

\addtolength{\textheight}{-0.5cm}

\bibliographystyle{ieee}
\bibliography{egbib}

\end{document}